\documentclass[runningheads]{llncs}
\usepackage[T1]{fontenc}

\usepackage{graphicx}
\usepackage[breaklinks=true]{hyperref}
\usepackage{subcaption}
\usepackage{array}
\usepackage{longtable}
\usepackage[inline]{enumitem}
\usepackage{listings}
\usepackage{caption}
\usepackage{amsmath}
\usepackage{cleveref}
\usepackage{makecell}
\usepackage{multirow}
\usepackage{wrapfig}
\usepackage[table]{xcolor}
\definecolor{KTOBlue}{RGB}{198,250,200}
\definecolor{KTOBlue2}{RGB}{230,200,239}
\definecolor{KTOBlue3}{RGB}{198,219,239}
\definecolor{SOTAOrange}{RGB}{254,230,206}
\usepackage[most]{tcolorbox}
\tcbuselibrary{listingsutf8}

\usepackage{enumitem}
\usepackage{parskip} %
\setlist{nosep} %

\tcbset{examplestyle/.style={
  listing only,
  breakable,
  colback=gray!5,
  colframe=gray!40!black,
  listing options={
    basicstyle=\ttfamily\footnotesize,
    breaklines=true,
    breakatwhitespace=true,
    columns=fullflexible
  }
}}

\definecolor{codegreen}{rgb}{0,0.6,0}
\definecolor{codegray}{rgb}{0.5,0.5,0.5}
\definecolor{codepurple}{rgb}{0.58,0,0.82}
\definecolor{backcolour}{rgb}{0.95,0.95,0.92}

\usepackage{algorithm}
\usepackage{algorithmic}
\usepackage{pdfpages}

\usepackage{longtable}%
\usepackage{svg}
\usepackage{enumitem}
\setlist[itemize]{noitemsep}
\setlist[enumerate]{noitemsep}
\setlist[enumerate]{topsep=2pt, itemsep=1pt, parsep=0pt}
\usepackage{booktabs}
\usepackage[load-configurations=version-1]{siunitx} %

\usepackage[disable]{todonotes}
\newcommand{\oshani}[1]{\todo[inline, color=blue!40, author=\textbf{Oshani}]{#1}}
\newcommand{\fernando}[1]{\todo[inline, color=red!40, author=\textbf{Fernando}]{#1}}

\begin{document}
\title{Federated Personal Knowledge Graph Completion with Lightweight Large Language Models for Personalized Recommendations}

\titlerunning{FedTREK-LM}

\author{Fernando Spadea\orcidID{0009-0006-4278-3666} \and
Oshani Seneviratne\orcidID{0000-0001-8518-917X}}

\authorrunning{Fernando Spadea and Oshani Seneviratne}

\institute{Rensselaer Polytechnic Institute, Troy, NY, USA \\
\email{\{spadef, senevo\}@rpi.edu}}

\maketitle              %
\begin{abstract}
Personalized recommendation increasingly relies on private user data, motivating approaches that can adapt to individuals without centralizing their information. We present Federated Targeted Recommendations with Evolving Knowledge graphs and Language Models (FedTREK-LM), a framework that unifies lightweight large language models (LLMs), evolving personal knowledge graphs (PKGs), federated learning (FL), and Kahneman-Tversky Optimization to enable scalable, decentralized personalization. By prompting LLMs with structured PKGs, FedTREK-LM performs context-aware reasoning for personalized recommendation tasks such as movie and recipe suggestions. 
Across three lightweight Qwen3 models (0.6B, 1.7B, 4B), FedTREK-LM consistently and substantially outperforms state-of-the-art KG completion and federated recommendation baselines (HAKE, KBGAT, and FedKGRec), achieving more than a $4\times$ improvement in F1-score on the movie and food benchmarks.
\oshani{4× for a higher F1-score, right? Need to make that explicit.}
\fernando{reworded this to "even achieving an F1-score over four times higher"}
Our results further show that real user data is critical for effective personalization, as synthetic data degrades performance by up to 46\%. Overall, FedTREK-LM offers a practical paradigm for adaptive, LLM-powered personalization that generalizes across decentralized, evolving user PKGs.

\keywords{
Knowledge Graph \and
Federated Learning \and
Knowledge Graph Completion \and
Personal Knowledge Graph \and
Kahneman-Tversky Optimization \and
Fine-Tuning \and
Large Language Model
}

\end{abstract}
\section{Introduction}
\label{sec:intro}

Delivering high-quality personalized recommendations, whether for movies, medications, or financial advice, requires models that adapt to individual user preferences. 
As user interactions increasingly span conversational systems, mobile devices, and federated platforms, building adaptive and privacy-preserving user models has become a central challenge in personalization research.
Traditional recommender systems struggle to achieve this balance, especially when personalization relies on structured user data such as personal knowledge graphs (PKGs).

While traditional KG completion methods~\cite{nathani2019learning,zhang2020learning,ma2024fedkgrec} have made progress for general recommendation tasks, they fall short for personalized recommendations with PKGs. 
These PKGs are inherently private and sparse as they are constructed and updated through ongoing user interactions, such as conversation history, app usage, or preference selections that reflect evolving user interests and contexts, meaning that these models must be able to adapt to these constantly evolving PKGs. 
For example, a movie or recipe recommender must continuously integrate feedback from new user interactions while maintaining the coherence of past preferences, which requires continual, preferably on-device, model adaptation.
As a result, such centralized models struggle to generalize across user-specific data distributions and adjust to evolving PKGs.

Moreover, personalization at the web-scale increasingly requires approaches that can learn collaboratively across users, leveraging shared behavioral signals, while respecting local data sovereignty~\cite{ibanez2023trust}. This motivates a federated, socially informed, learning paradigm for user modeling and recommendation. Additionally, LLMs are trained on vast swaths of public data, but they cannot access private, personal data on user devices. Federation enables the use of this invaluable data without violating the users' sovereignty over their personal information. Accessing this data is becoming increasingly important, as alternatives, such as synthetic data generated by AI, have been shown to be ineffective~\cite{shumailov2024ai}.

In this paper, we introduce Federated Targeted Recommendations with Evolving Knowledge graphs and Language Models (\textbf{FedTREK-LM}), a novel framework that combines federated learning (FL)~\cite{mcmahan2017communication}, lightweight LLMs, and feedback-driven fine-tuning to perform personalized recommendation. FedTREK-LM operates in a decentralized manner: each client fine-tunes a small LLM locally using Kahneman-Tversky Optimization (KTO)~\cite{ethayarajh2024kto}, based on user interactions encapsulated in their PKG. These model updates are aggregated by a central server, allowing the global model to improve without accessing raw user data. 
We incorporate lightweight LLMs because they offer strong generalization and reasoning capabilities, and prompting them with structured PKGs enables context-specific recommendations~\cite{zhang2024making}. However, fine-tuning LLMs directly on user devices is challenging due to constrained computational resources (i.e., limited memory and processing power) and due to the sparsity of supervision signals (i.e., the limited availability of high-quality labeled examples or explicit user feedback). KTO addresses both issues: it enables lightweight, label-efficient fine-tuning from user feedback, and it supports continual learning over dynamic graphs.

\begin{figure*}[t]
    \centering
    \includegraphics[width=\textwidth, alt={A composite graphic containing a scatter plot and four bar charts. The scatter plot graphs Precision on the x-axis versus Recall on the y-axis for various models including TREK-LM, KBGAT, HAKE, and FedKGRec across centralized and federated settings. The TREK-LM models are clustered in the top-right, indicating higher overall performance. The four bar charts compare F1-Score, MRR, Hits@1, and Hits@10, illustrating that TREK-LM at 0.6B, 1.7B, and 4B parameter sizes consistently outperforms the baseline models in all evaluated metrics.}]{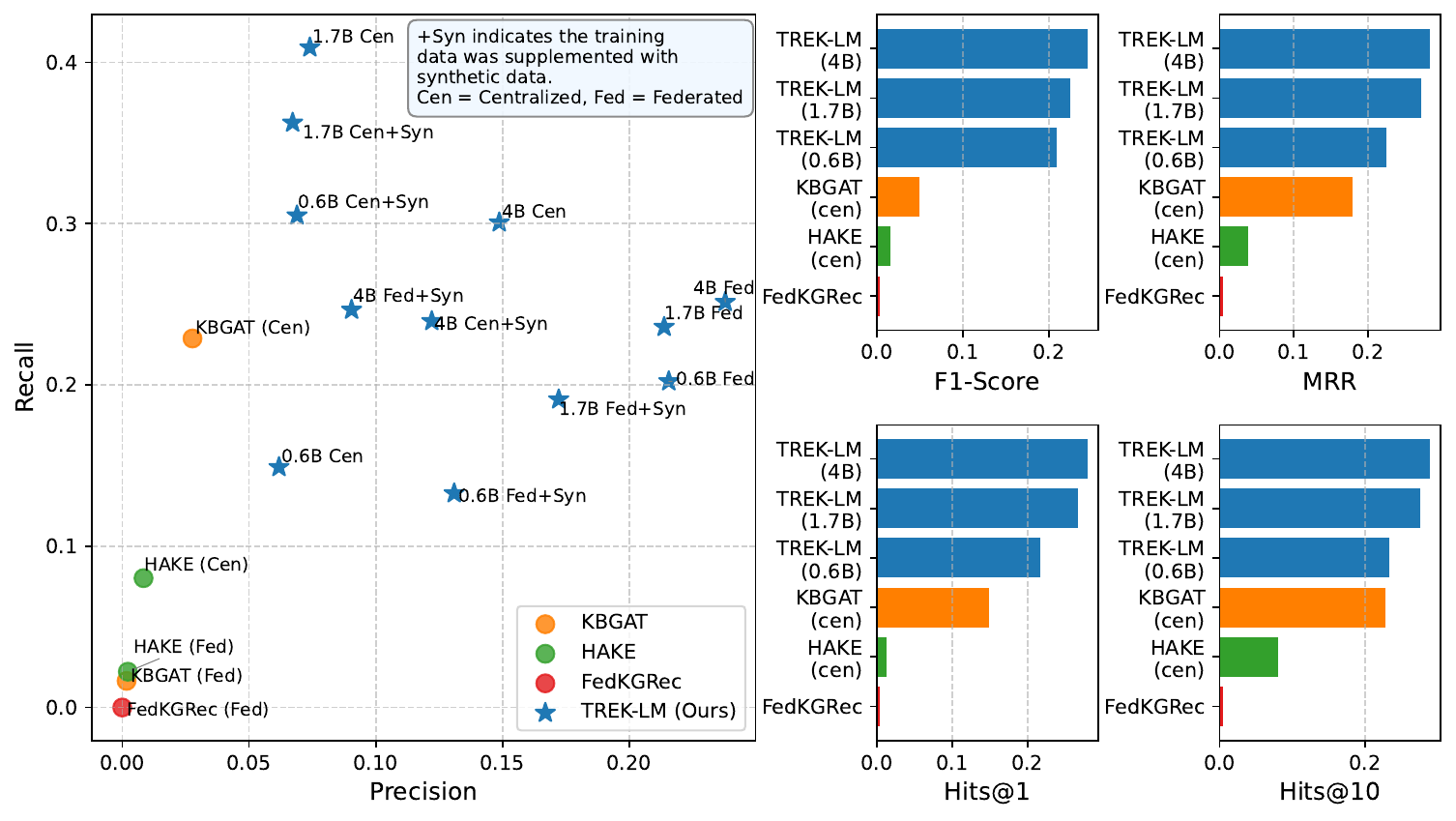}
    \caption{
    Overall performance results of our TREK-LM models against the baselines (KBGAT, HAKE, and FedKGRec).
    }
    \label{fig:hook}
\end{figure*}

\paragraph{Contributions:} 
\begin{enumerate}
\item We develop FedTREK-LM, an FL pipeline for lightweight LLM-based, personalized recommendations, that operates over evolving PKGs encoded in JSON-LD, ensuring compatibility with Semantic Web standards and decentralized data settings.
\item We introduce a feedback-driven, lightweight fine-tuning strategy based on KTO that enables efficient on-device adaptation using real interactions, supporting continual learning over dynamic, user-specific graphs while remaining computationally feasible.
\item We empirically show that federated access to real user data is important, as synthetic data substantially degrades performance in both centralized and federated settings, highlighting the limitations of synthetic conversational data for PKG construction and personalized reasoning.
\item We show how natural language can guide LLMs to reason over PKGs for targeted recommendations across domains such as movies and recipes, demonstrating that structured PKG prompting enables context-aware semantic reasoning in lightweight LLMs.
\item We show that FedTREK-LM outperforms state-of-the-art KG completion models such as KBGAT~\cite{nathani2019learning} and HAKE~\cite{zhang2020learning} in both centralized and federated settings. Additionally, FedTREK-LM outperforms FedKGRec~\cite{ma2024fedkgrec}. An overview of these results can be seen in \Cref{fig:hook}, where we refer to the base LLM component of our framework as TREK-LM.
\item We show that FedTREK-LM significantly outperforms purely local models, demonstrating that federated collaboration is essential for overcoming data sparsity in PKGs and enabling robust personalization.
\end{enumerate}

\vspace{-2mm}

\section{Related Work}
\label{sec:relwork}

\vspace{-1mm}

Our work bridges three major research areas: KG completion for recommendations, personalized recommendation with LLMs, and federated model adaptation. 
While each area has made great progress individually, existing approaches struggle to support evolving PKGs and decentralized, interaction-driven model updates, highlighting the need for an integrated solution like FedTREK-LM.

\vspace{-2mm}

\paragraph{Recommendation using KG Completion in Centralized Settings.}
Traditional KG completion methods, such as KBGAT~\cite{nathani2019learning} and HAKE~\cite{zhang2020learning}, use graph embeddings to infer missing links between entities. Specifically, KBGAT uses attention mechanisms to weigh neighborhood relevance, while HAKE encodes hierarchical structure using polar embeddings. While effective in static, centralized KG environments, these models lack support for personalization, cannot adapt to evolving user graphs, and require full KG access, making them unsuitable for decentralized, user-specific applications.
Recent work has explored integrating user modeling within KG-based recommenders~\cite{wang2019kgat,xian2019reinforcement}, highlighting the benefits of user-aware embeddings and attention mechanisms, but these remain primarily centralized and non-interactive.
Moreover, nearly all of these approaches assume a single global KG and do not support the construction or maintenance of \emph{PKGs} that evolve over time.

\vspace{-2mm}

\paragraph{LLMs for Personalized Recommendations.}
Recent studies explore using LLMs to extract user preferences and populate structured KGs from conversational data~\cite{qiu2024unveiling}. Others have shown that general-purpose LLMs like ChatGPT can complete KGs using zero-shot prompts~\cite{meyer2023llm}. 
Several works demonstrate the value of PKGs for health-related personalization~\cite{seneviratne2021personal,shirai2021applying,seneviratne2023semantically,yang2024transforming}.
However, these approaches primarily operate over static or centrally maintained KGs and lack mechanisms for continual, feedback-driven updates to PKGs on user devices.
In contrast, FedTREK-LM performs real-time, interaction-driven KG completion by fine-tuning lightweight LLMs on evolving PKGs.

\vspace{-2mm}

\paragraph{Federated Model Adaptation.}
In the recommender systems literature, personalization is often achieved via collaborative filtering~\cite{su2009survey}, but these methods typically assume centralized access to user data. In contrast, real-world applications increasingly involve decentralized, device-local data, such as viewing histories, interactions, or evolving PKGs, distributed across user devices. Few works address LLM-based recommendation in such settings, but those that do, focus on highly specialized applications rather than general solutions~\cite{long2023decentralized,chen2022gdsrec}. Our approach fills this gap by demonstrating that federated fine-tuning of lightweight LLMs using KTO enables effective personalization over decentralized data without requiring raw data to be centralized.
Additionally, FedE~\cite{chen2021fede}, and FedRecKG~\cite{ma2024fedkgrec} propose embedding-based models trained in decentralized environments. However, these systems still rely on global KGs or flat user–item graphs, offer limited user-level personalization, do not use LLMs, and do not support evolving KGs with new entities during training.
Our framework differs by operating over natural language prompts, using PKGs, and adapting dynamically to interaction-derived feedback~\cite{spadea2025bursting,spadea2025avoiding,spadea2025enhancing}, providing the flexibility to be adapted to more domains~\cite{spadea2025aligning,spadea2025parallel}.

\begin{figure}[htbp]
    \centering
    \includegraphics[width=0.9\columnwidth, alt={An architectural diagram of the FedTREK-LM system. It displays a central server that performs Federated Aggregation on weight updates. The server is connected to multiple distinct clients, labeled Client 1 through Client N. On each client device, a local Large Language Model and a Personal Knowledge Graph interact alongside KTO Fine-Tuning to generate unique model updates, which are then transmitted back to the central server.}]{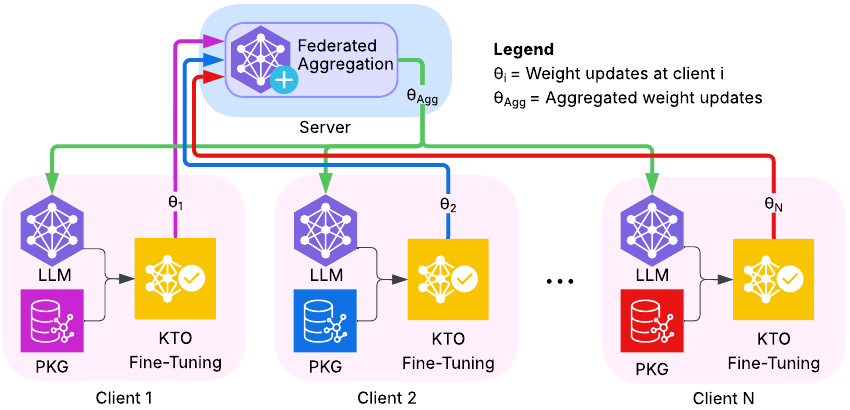}
    \caption{
    Overview of the FedTREK-LM architecture.
    }
    \label{fig:fed}
\end{figure}

\section{System Architecture}
\label{sec:modes}

We developed a model architecture (see \Cref{fig:fed}) centered around a lightweight LLM fine-tuned in an FL setting to make user recommendations using their PKG. 
In this architecture, each user operates as an FL client, fine-tuning a local model and uploading their model updates to the central server without exposing their private data. That way, the server is able to aggregate the model updates to create a new global model and send the new updated model to all the clients.

\begin{figure}[t]
    \centering
    \begin{subfigure}[b]{0.48\textwidth}
        \centering
        \includegraphics[width=\textwidth, alt={flowchart titled "Client-Side KTO Training" detailing the local feedback-driven fine-tuning process. It shows KTO Training Data providing a Prompt, a Label, and a Completion to a KTO Loss Function. The Prompt is also sent to an LLM, which outputs an LLM Completion to the KTO Loss Function. The KTO Loss Function then sends Model Updates back to the LLM.}]{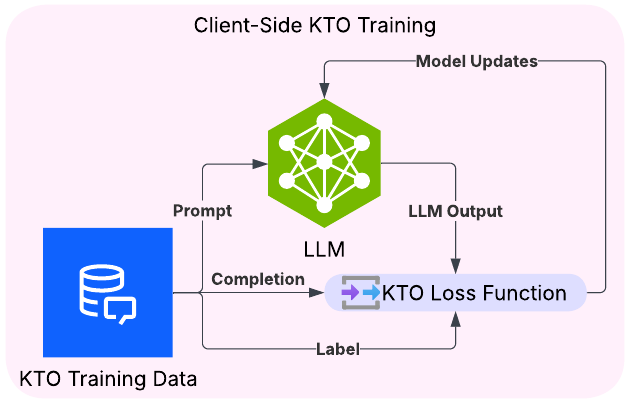}
        \caption{Client-side KTO training process.}
        \label{fig:train}
    \end{subfigure}
    \hfill
    \begin{subfigure}[b]{0.48\textwidth}
        \centering
        \includegraphics[width=\textwidth, alt={A flowchart titled "Client-Side Knowledge Graph Completion" detailing the local recommendation workflow and how the LLM reasons over the PKG. A User Request is sent to a PKG and an LLM. The PKG generates a Relevant Query and extracts a SubPKG, which is sent to the LLM. The LLM produces a Response to Derive Recommendation, which in turn generates a Knowledge Graph Update that feeds back into the PKG.}]{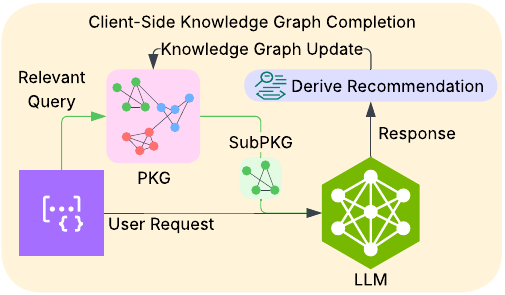}
        \caption{Client-side recommendation workflow.}
        \label{fig:kgc}
    \end{subfigure}
    \caption{Detailed view of the FedTREK-LM local operations: (a) shows the feedback-driven fine-tuning using KTO, and (b) illustrates how the LLM reasons over the PKG to derive recommendations.}
    \label{fig:local_ops}
\end{figure}

The local model is fine-tuned using KTO~\cite{ethayarajh2024kto}, a feedback-driven optimization strategy that supports lightweight, flexible supervision. As seen in \Cref{fig:train}, prompt-completion-label triples from the user’s PKG interaction history are used to compute the KTO loss, which fine-tunes the LLM locally. The resulting model updates are then sent to the server for federated aggregation.
KTO enables the integration of user interactions, facilitating continual personalization even in sparse and decentralized data regimes.

\subsection{Structured Prompting with PKGs}
\label{subsec:promptDesign}

Each client maintains a PKG that grows over time as the user interacts with the LLM. As shown in \Cref{fig:kgc}, when the user poses a query, the system constructs a prompt for the LLM by querying the user's PKG for a relevant subPKG (i.e., a subgraph with the user's movie preferences in the case of a movie recommendation application) that is then prepended to the request via a system message that defines the model's role using the format given in \Cref{ex:sys} in \ref{sec:prompts}. 
\oshani{What's exactly in this query to select the SubPKG? Since we got some questions in the WebConf reviews, it would be good to make it clearer here.}
\fernando{Added (i.e., a subgraph with the user's movie preferences for a movie recommendation application)}

\subsection{Knowledge Representation}

\oshani{We should show a small JSON-LD snippet with @context and explicit IRIs, possibly in the appendix, and refer to it from here.}
\oshani{I also think it would be good to clarify how users and items are typed, i.e., that it's mappable to schema.org, maybe?}
\fernando{Not sure what you're getting at with the typing stuff, but I added this: Users and items are typed using standards adhearing to schema.org to make the PKG easily queryable.}
\oshani{Because this a semantic web conference, and the reviewers will care about this stuff.}
\fernando{Yeah, I understand that. I just don't know what it means.}

Each client maintains a PKG represented in JSON-LD %
that encodes user-specific preferences (e.g., liked or disliked entities) as a structured, semantically grounded graph. This representation supports explicit entity typing and relation modeling, enabling both machine interpretability and compatibility with downstream reasoning.
Users and items are annotated with schema.org classes and properties, ensuring interoperability with existing vocabularies and allowing the PKG to be queried, extended, and integrated using standard Semantic Web practices.
\oshani{Let's be clear about what counts as positive / negative here. In a way, this ties to the "Human in the loop" aspect of the CFP, so would be advantageous for us.}
\fernando{I just commented out the references to this stuff because we don't need to explain these things before we get into KTO in 3.4.}

\subsection{Federated Training Process}

Model training proceeds in communication rounds, each consisting of the following steps:

\begin{enumerate}[topsep=2pt, itemsep=2pt, parsep=0pt]
    \item The current global model is sent to a selected subset of clients.
    \item Each client fine-tunes the model on their locally maintained PKG using KTO, using real user interactions to update the model’s ability to complete and extend their PKG. 
    \item The locally updated models, represented as lightweight LoRA updates~\cite{hu2021lora}, are sent back to the central server.
    \item The server aggregates these updates (e.g., via Federated Averaging~\cite{mcmahan2017communication}) to produce a new global model that captures knowledge acquisition patterns across diverse users while preserving their data locality.
\end{enumerate}

This process repeats iteratively, enabling the global model to learn PKG recommendation strategies from decentralized user data. Over time, the model captures generalizable patterns of PKG evolution, such as user preference shifts or emerging entity associations.

One of the primary benefits of the federated setup is its ability to support cross-user generalization. Even though user data never leaves their local data store, the aggregation of model updates enables the global model to learn patterns that generalize across users with similar preferences. This leads to more effective PKG recommendations for each user by leveraging the collective interactions of the broader user base. For example, if multiple users who enjoy suspense thrillers respond positively to \texttt{Zodiac (2007)}, the global model may recommend \texttt{Zodiac} to another user with similar latent preferences, even if that user has never interacted with content in the thriller genre before. This blend of local adaptation and global generalization enables the model to learn beyond any single user’s data distribution and to generalize to novel entities and preferences.

\subsection{KTO Application}
\label{sec:KTOApp}

\begin{figure}[htbp]
    \centering
    \includegraphics[width=0.9\columnwidth, alt={A block diagram breaking down the Client-Side KTO Data Construction process. A chat exchange is divided into three distinct functional parts: a System Prompt with the Knowledge Graph and the User Request combine to form the Prompt; the Model Response forms the Completion; and User Feedback is used to derive the Label. Together, these three compiled elements constitute the KTO Training Data.}]{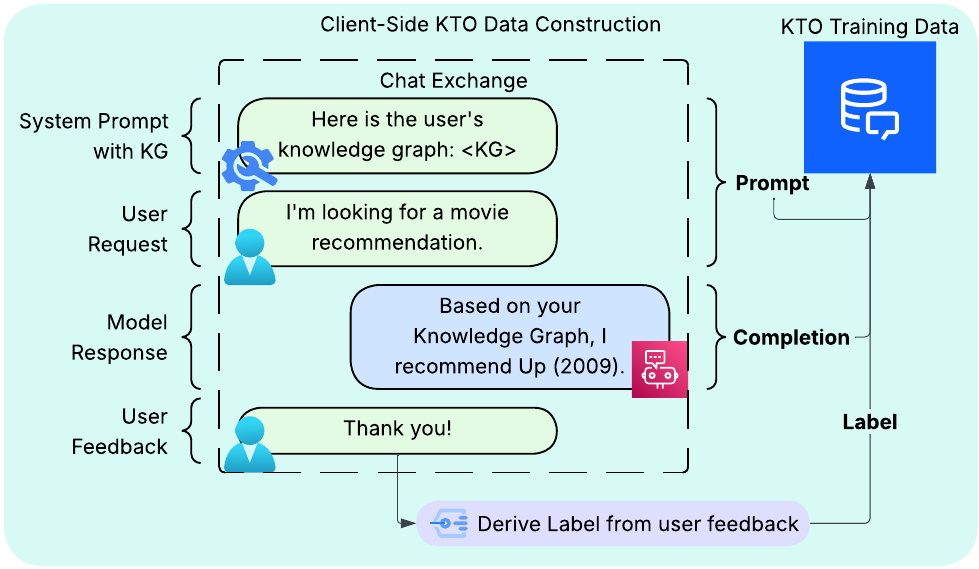}
    \vspace{-2mm}
    \caption{Illustration of KTO data construction in the form of prompt–completion–label triplets from user-LLM interactions. 
    }
    \label{fig:data}
\end{figure}

To enable lightweight and scalable fine-tuning of LLMs for PKG recommendation, we apply KTO~\cite{ethayarajh2024kto} locally on clients throughout each round of federated training.
KTO is particularly well-suited to this setting, as it avoids the need for fully supervised labels or token-level annotations. 
Recent comparative work~\cite{spadea2025federated,saeidi2024insights} demonstrates that KTO significantly outperforms other preference-based fine-tuning algorithms, making it a strong choice for real-world, decentralized personalization tasks like ours.
To support preference-based fine-tuning in our framework, each user interaction is distilled into a simple training instance consisting of three components: \emph{prompt–completion–label} triplets, as illustrated in \Cref{fig:data}. An example for how these three components are constructed is provided in \Cref{subsec:dataset}.
\begin{enumerate}[label=(\arabic*)]
\item \textbf{Prompt}: Natural language input or a truncated user–LLM conversation.
\item \textbf{Completion}: A potential response to the prompt.
\item \textbf{Label}: A binary signal indicating whether the completion is desired or not.
\end{enumerate}

These triplets are generated continuously as users interact with the LLM. 
Feedback, either explicit or implicit, is used to label completions.

While KTO is a flexible mechanism for fine-tuning LLMs using feedback-driven supervision, it relies on a reference model to compute the alignment loss between the model’s current and prior outputs. In a naive implementation, this would require storing both a trainable and a frozen copy of the model in memory, effectively doubling the memory footprint. This is particularly prohibitive in our federated setting, where clients would want to run the model on memory-constrained devices such as smartphones.
To address this, we adopt LoRA~\cite{hu2021lora} via the PEFT framework~\cite{pu2023empirical}. Because LoRA freezes the base model and introduces a small number of trainable low-rank matrices, the frozen model can also serve as the reference in KTO’s loss computation~\cite{ye2024openfedllm}, thereby avoiding the need to store two distinct model copies.
This memory efficiency is particularly important on low-resource consumer devices. 
LoRA enables efficient adaptation to new triples or domains with minimal parameter updates, ensuring that fine-tuning remains computationally feasible. Furthermore, the reduced number of trainable parameters leads to significantly smaller communication payloads in the federated setting, lowering bandwidth costs during model aggregation and improving scalability across users with heterogeneous KGs.
Beyond computational and communication efficiencies, the use of LoRA also acts as an inherent partial mitigation against privacy risks. While federated learning is a privacy-enhancing technology rather than a strict privacy guarantee, transmitting only low-rank adapter matrices rather than full model gradients reduces the attack surface for model inversion~\cite{bossy2025mitigating}. We intentionally did not implement a formal privacy guarantee, such as differential privacy, in this study to avoid confounding our utility results, though integrating such mechanisms remains a valuable direction for future work.

\section{Experimental Setup and Training Protocol}
\label{sec:method}

\oshani{This first paragraph can go to an appendix to save space for semantic/KG details.}
\fernando{I moved the first two sentences to "Hardware and Software Specifications" in the appendix.}
\oshani{Additionally, you probably should move the Lora hyperparameters, number of epochs, optimization details of our work as well as the competitors' implementation for completeness.}
\fernando{I think I've done this, although I don't know what some of the things you were referring to meant. I also realized that the mention of 128 in the communication costs section came out of nowhere since it wasn't mentioned before, so I explained that better.}
\oshani{I meant if you configured anything specific in KBGAT, HAKE, and FedKGRec (our competitors) those configs should also be in the appendix.}
\fernando{addressed now}
In our framework, a user's PKG is provided to the LLM as part of the input prompt during a conversation. The LLM is instructed, via a system-level prompt, to use this PKG to inform its recommendations. The goal is for the LLM to infer or suggest new, relevant entities that the user would like.

\subsection{Dataset Construction}
\label{subsec:dataset}

We base our experiments on the following two derived datasets.

\begin{itemize}[leftmargin=*,label={}]
    \item \textbf{\texttt{Movie PKG}:} This is based on the \textit{Recommendation Dialogues} (ReDial)~\cite{li2018towards,huggingface2024redial} dataset, which contains conversations between two users (an initiator and a respondent) discussing movie preferences. ReDial is well-suited for our setting because it provides explicit, temporally ordered user interactions and captures evolving preference signals that naturally emerge over the course of dialogue.
    To the best of our knowledge, ReDial is currently the only publicly available recommendation-based conversation dataset that includes annotated metadata for both recommendations and sentiment.
    \item \textbf{\texttt{Recipe PKG}:} We used the \textit{Food.com Recipes and Interactions}~\cite{shuyang_li_2019} dataset, which includes recipes annotated with structured metadata (e.g., ingredients, tags) and user ratings from 0 to 5 stars.
    Although it does not include conversational data like ReDial, we generate synthetic data points from two structured KGs that we construct:
    (1) \textit{Recipe KGs}, encoding recipe-level semantics such as ingredients and tags; and (2) \textit{User KGs}, capturing user ratings for recipes.
\end{itemize}

To simulate an FL environment, we partition the Movie PKG dataset by conversation initiator and Recipe PKG dataset by the Food.com user. Each initiator/user is treated as a distinct client, and all conversations they start are grouped to form their local dataset. This design simulates a realistic FL scenario where each client maintains its own evolving PKG, built from the user’s prior and evolving likes and dislikes.

\subsubsection{Real Movie Data Construction.}

For the \texttt{Movie PKG} dataset, each training example consists of a prompt, completion, and label determined by: 
\begin{enumerate}
    \item \textbf{Prompt}: Includes a truncated conversation up to a point where a movie recommendation is made by the respondent. It also includes the user's PKG up to that point, constructed chronologically by collecting all stated movie preferences from earlier conversations.
    \item \textbf{Completion}: Corresponds to a recommendation, typically a list of movies, derived from the movies that were actually recommended in the respondent's message. 
    \item \textbf{Label}: Binary signal indicating where the recommendations are desired. If they were disliked or if the recommended movies were already present in the PKG (indicating redundancy), it is labeled as negative. Otherwise, it is labeled as positive.
\end{enumerate}

If a conversation contains multiple movie recommendation events, we extract multiple training examples. Similarly, if a single message contains both good and bad suggestions, we separate them into separate positive and negative examples.

\vspace{-4mm}

\subsubsection{Synthetic Data Generation.}

Since the Food.com dataset does not include any conversations, the \texttt{Recipe PKG} data points need to be generated. Additionally, to enhance data diversity and reduce overfitting, we add \textit{synthetic KTO examples} alongside the real examples of the \texttt{Movie PKG} dataset. The synthetic data points for both the \texttt{Recipe PKG} and \texttt{Movie PKG} datasets are generated through the following two techniques:

\begin{enumerate}[topsep=2pt, itemsep=2pt, parsep=0pt, label=(\roman*)]
    \item \textbf{Masking}, where random triples from the PKG are hidden and used as the completion to train the model to recover or avoid them depending on whether the triples represented positive or negative preferences.
    \item \textbf{Penalizing Redundancy}, where existing triples from the KG are used as 
    negative examples to discourage the model from generating completions that simply repeat facts already present in the PKG.
\end{enumerate}

We evaluate the model’s performance on the \texttt{Movie PKG} dataset both with and without synthetic data to isolate its contribution. %
In total, the base \texttt{Movie PKG} dataset contains \emph{34,800 real data points across 355 clients}, while the version with the synthetic data includes \emph{67,922 total data points across 512 clients}. An additional \emph{3,603 real data points} are held out for testing. The \texttt{Recipe PKG} dataset ends up with \emph{25,236 data points across 940 clients} with \emph{3,000 data points} held for testing. Prompt–completion pair examples are shown in \ref{sec:pce} for real (\Cref{ex:real}) and synthetic (\Cref{ex:syn}) user interactions. Because the data is split by real user data, it is inherently \textbf{not} independent and identically distributed. Unlike a uniform split, this means different clients possess vastly different counts of data points, reflecting natural, uneven usage patterns.

\subsection{Model Selection}

For our experiments, we selected the Qwen3~\cite{huggingface2024qwen3} family of models, specifically the smallest 0.6B, 1.7B, and 4B sizes. This smaller-scale model was deliberately chosen to reflect the practical constraints (limited compute and memory) of deploying LLMs in a federated setting on edge devices~\cite{githubExecutorchexamplesmodelsqwen3READMEmdMain}. 
Larger models are infeasible to run, much less fine-tune, on such devices. In contrast, Qwen3-0.6B is small enough to be trained on typical consumer-level hardware, making it realistic for use in FL \cite{githubExecutorchexamplesmodelsqwen3READMEmdMain}.
To further reduce the computational and memory overhead, we applied 4-bit quantization to the model weights. This significantly reduces the size of the model and the memory required for training and inference, while still maintaining acceptable performance. Quantization is especially useful in our federated setup because it lowers training time and communication costs when transferring model updates between clients and the central server.

\vspace{-3mm}

\subsection{Communication Cost}

\begin{table}
\centering
\caption{Communication costs for FedTREK-LM models between the individual client overhead per active round and the total aggregated server load.}
\label{tab:comm_costs}
\begin{tabular}{lccc}
\toprule
\textbf{Model} & 
\makecell[c]{\textbf{Trainable} \\ \textbf{Parameters}} & 
\makecell[c]{\textbf{Per-Round} \\ \textbf{Client Cost}} & 
\makecell[c]{\textbf{128-Round Total} \\ \textbf{Server Cost}} \\
\midrule
Qwen3-0.6B & 10.093 M & 38.50 MB & 38.5 GB \\
Qwen3-1.7B & 17.433 M & 66.50 MB & 66.5 GB \\
Qwen3-4B   & 33.030 M & 126.00 MB & 126.0 GB \\
\bottomrule
\end{tabular}
\end{table}

In our federated learning setup, communication costs depend entirely on the size of the transmitted model weights. Because we use parameter-efficient fine-tuning via LoRA, clients only need to exchange compact adapter weights with the server, rather than the full model. In each training round, four selected clients fine-tune the model locally and upload their updated LoRA adapters. The central server aggregates these updates and broadcasts the new global model to four different clients for the next round.

To properly contextualize these costs, it helps to distinguish the minimal bandwidth required from an individual client from the cumulative load handled by the central server. The \textbf{Client Cost (Per Round)} reflects the one-way bandwidth a client uses to either download or upload the LoRA adapter. As \Cref{tab:comm_costs} shows, which reports adapter parameter counts and communication costs, this client-side cost ranges from just 38.50 MB to 126.00 MB. This is roughly equivalent to streaming a short video and easily fits within the limits of standard Wi-Fi or 4G/5G mobile networks.

On the other hand, the \textbf{Server Cost (128 Rounds)} represents the total aggregated data processed by the central server over the entire training lifecycle (assuming a LoRA rank of 16 and 32-bit parameters). We calculate this by taking the one-way client cost, doubling it to account for both the download and upload phases, and multiplying that by the four clients participating in each of the 128 total rounds. Thus, the 126.0 GB total for the Qwen3-4B model is strictly a server-side cumulative cost, not a burden placed on any single edge device.

\subsection{Evaluation Approach}

To assess our model’s ability to perform effective KG completion, we use a held-out set of positive data points drawn from the two PKG datasets. For each prompt, we evaluate whether the model generates accurate recommendations by comparing its output to the reference completions. We compute true positives (TP), false positives (FP), and false negatives (FN), where a TP is a recommendation that appears in the ground truth, an FP is an incorrect or irrelevant recommendation, and an FN is a ground truth item that the model fails to recommend. From these counts, we compute precision ($\text{TP} / (\text{TP} + \text{FP})$) and recall ($\text{TP} / (\text{TP} + \text{FN})$), capturing the accuracy and coverage of the model’s predictions, respectively. With these values, we then also calculate the F1-scores ($2/\left(\frac{1}{precision} + \frac{1}{recall}\right)$).
\oshani{Is the F1 score correct? See: https://en.wikipedia.org/wiki/F-score. Regardless, do we need these definitions for precision, recall, and F1 score, since they are standard metrics?}

In addition to classification-based metrics, we evaluate ranking performance using Hits@1, Hits@3, Hits@10, and Mean Reciprocal Rank (MRR). These metrics are computed based on the position of the first correct recommendation in the model's output. If the first recommendation is correct, the rank is 1; if it is the second, the rank is 2; and so on. If no correct prediction appears, we assign reciprocal rank 0 (i.e., treat rank as $\infty$).
Hits@k measures the fraction of queries for which a correct recommendation appears within the top k predictions, while MRR averages the reciprocals of the ranks.
These metrics reflect how early relevant recommendations appear in the output, which is critical for practical systems where top-ranked results are most visible.

Additionally, we perform several ablations:

\begin{enumerate}
    \item \textbf{Synthetic vs Real:} We train the models with and without synthetic data.
    \item \textbf{Centralized vs Federated:} We consider the centralized setting in addition to the federated setting to measure the performance degradation in FL.
    \item \textbf{Local Training Only:} We test models trained on a single user's data to demonstrate that FL is an improvement over purely local models. We train 10 models on 10 random users (out of those with at least 10 data points for evaluation) and take the average of their results.
    \item \textbf{KTO Fine-Tuned vs Base:} Furthermore, we compared our KTO fine-tuned FedTREK-LM models against the corresponding base Qwen3 models to demonstrate that KTO fine-tuning improves the results.
    \item \textbf{Model Size:} We compare the results of the 0.6B, 1.7B, and 4B Qwen3 models to evaluate the significance of the model size on the performance.
\end{enumerate}

To benchmark our method against traditional KG completion models for recommendations, we compare it to HAKE~\cite{zhang2020learning} and KBGAT~\cite{nathani2019learning}, which we adapt to operate under FL constraints. We chose these two models because they were the top performers on the \textit{paperswithcode} leaderboards for the KG completion task on several datasets~\cite{kbgathakebench}. While these models perform well in centralized benchmarks, they assume a complete, static dictionary of entities and relations, a requirement incompatible with FL's decentralized nature. Moreover, they are inflexible to dynamic KGs, as they cannot accommodate new entities after initialization, so they need to restart training when new entities are introduced. This would be a major problem in a federated setting, as the model may no longer have access to data it originally trained on if some clients leave the network. 
We also benchmark against FedKGRec~\cite{ma2024fedkgrec}, which is specifically designed for federated recommendations. However, FedKGRec, HAKE, and KBGAT are not designed to work with PKGs. Therefore, to ensure fair comparison, we flatten each user's KG into a single list of triples and construct a centralized training set that mimics the train/test split of our KTO datasets used for evaluating FedTREK-LM. For KBGAT, which uses a graph attention network (GAT) followed by a convolution-based prediction module, we implement a federated version by first training the GAT locally on each client and then training the convolutional KG completion model via FL.

\section{Results}
\label{sec:results}

\begin{table*}[hbt]
\centering
\caption{Comparative performance on the Movie PKG recommendation task. We compare our TREK-LM variants (using the Qwen3-4B model) across centralized, federated, and local-only settings, with and without synthetic data (\textit{+Syn}). Baselines include traditional KG completion models (KBGAT and HAKE) and a federated recommendation model (FedKGRec). The best result in each column is bolded.}
\label{tab:movie}
\resizebox{\textwidth}{!}{%
\begin{tabular}{llcccccccc}
\toprule
\textbf{Model} & \textbf{Setting} & \textbf{Precision} & \textbf{Recall} & \textbf{F1-score} & \textbf{MRR} & \textbf{Hits@1} & \textbf{Hits@3} & \textbf{Hits@10} \\
\midrule
\multirow{6.6}{*}{TREK-LM} 
 & Centralized & 0.149 & \textbf{0.301} & 0.199 & 0.247 & 0.192 & \textbf{0.299} & \textbf{0.321} \\
 & Centralized \textit{+Syn} & 0.122 & 0.240 & 0.162 & 0.209 & 0.177 & 0.238 & 0.255 \\
 \cmidrule{2-9}
 & \textbf{Federated} & \textbf{0.238} & 0.251 & \textbf{0.244} & \textbf{0.284} & \textbf{0.280} & 0.290 & 0.290 \\
 & Federated \textit{+Syn} & 0.090 & 0.247 & 0.132 & 0.218 & 0.183 & 0.254 & 0.270 \\
 \cmidrule{2-9}
 & Local-only & 0.015 & 0.068 & 0.025 & 0.069 & 0.056 & 0.081 & 0.081 \\
 & Local-only \textit{+Syn} & 0.015 & 0.068 & 0.025 & 0.069 & 0.056 & 0.081 & 0.081 \\
\midrule
\multirow{2}{*}{KBGAT} 
 & Centralized & 0.028 & 0.229 & 0.049 & 0.180 & 0.150 & 0.182 & 0.229 \\
 & Federated & 0.002 & 0.017 & 0.003 & 0.008 & 0.002 & 0.007 & 0.017 \\
\midrule
\multirow{2}{*}{HAKE} 
 & Centralized & 0.008 & 0.080 & 0.015 & 0.038 & 0.013 & 0.032 & 0.080 \\
 & Federated & 0.002 & 0.022 & 0.004 & 0.009 & 0.000 & 0.006 & 0.022 \\
\midrule
FedKGRec & Federated & 0.000 & 0.000 & 0.000 & 0.000 & 0.000 & 0.000 & 0.000 \\
\bottomrule
\end{tabular}
}
\end{table*}

\subsection{Movie PKG Results}
In \Cref{tab:movie}, we report recommendation quality metrics for the Movie PKG dataset across the baselines and our models trained in three settings: centralized, federated, and local-only, with (\texttt{+Syn}) and without synthetic data. The federated model without synthetic data achieves the best precision, F1, MRR, and Hits@1 scores, while the centralized model achieves the best results in recall, Hits@3, and Hits@10. KBGAT and HAKE perform poorly in both the federated and centralized settings. FedKGRec achieves the worst results as it fails to handle the dataset's sparsity.

\vspace{-2mm}

\subsubsection{Comparison with Baselines:}
Our model, TREK-LM, consistently outperforms the state-of-the-art KG completion models, KBGAT and HAKE, across all metrics and under both centralized and federated settings. The FedKGRec model could not handle the sparsity of the data, leading to zeros across the board. This highlights TREK-LM's ability to work well with sparse data.

\subsubsection{Centralized vs Federated:}
\begin{figure}[htbp]
    \centering
    \includegraphics[width=0.9\columnwidth, alt={A grouped bar chart comparing TREK-LM performance using the Qwen3-4B model across three training settings: Centralized without synthetic data, Federated without synthetic data, and Local-only. The metrics evaluated are Precision, Recall, F1-Score, MRR, and Hits@1. The federated setting scores visibly highest in Precision, F1-Score, MRR, and Hits@1, while the centralized setting scores highest in Recall. The Local-only performance bars are near zero across all metrics.}]{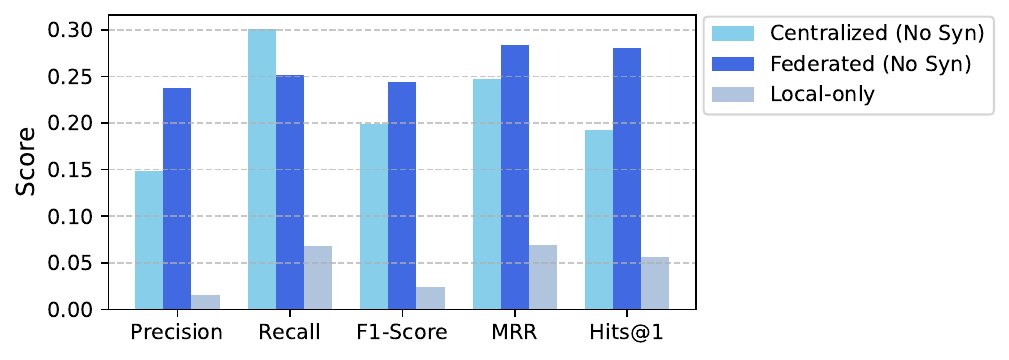}
    \vspace{-3mm}
    \caption{
    TREK-LM Performance by Setting (Qwen3-4B)
    }
    \label{fig:para}
\end{figure}

Notably, while HAKE and KBGAT suffer substantial performance degradation when adapted to an FL environment, FedTREK-LM models demonstrate strong resilience. KBGAT’s recall, for example, drops by 93\%, while HAKE’s drops by 73\%. In contrast, between centralized and federated models, our method's recall drops only 17\% without synthetic data and actually increases 3\% with it. Moreover, while recall decreases modestly under FL, the model achieves significant gains in precision, F1-score, MRR, and Hits@1 (see \Cref{fig:para} for a comparison across the various settings in our best-performing TREK-LM model, Qwen3-4B), suggesting that FL encouraged the model to prefer quality over quantity.

\subsubsection{Local Ablation Results:}
As can be seen in \Cref{fig:para}, the local models consistently underperform the federated models across all the metrics. As expected, the local model is not able to learn how to recommend new movies to the user because it has not been trained on any movies that the user has not yet interacted with. 
The few correct predictions are likely due to chance, as indicated by the identical results with and without synthetic data in \Cref{tab:movie}.
This further reinforces that learning from other users is extremely important for the model to be able to recommend new movies to the user.

\subsubsection{Effect of KTO Fine-Tuning:}
\begin{figure}[htbp]
    \centering
    \begin{minipage}[b]{0.42\textwidth}
        \centering
        \includegraphics[width=\textwidth, alt={A grouped bar chart comparing the F1-scores of Base Models against KTO Fine-Tuned Federated TREK-LM models across three parameter sizes: 0.6B, 1.7B, and 4B. For every model size category, the TREK-LM federated bar is significantly taller than its corresponding Base Model bar, demonstrating a clear performance improvement.}]{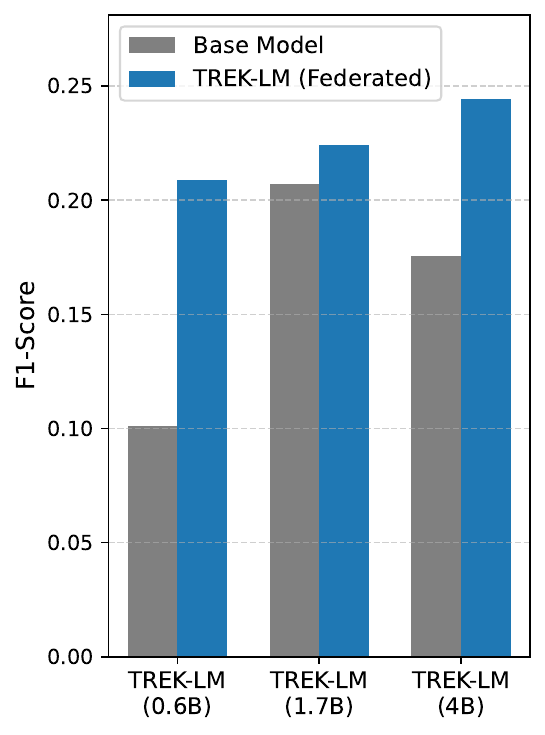}
        \caption{F1-scores of Base vs. KTO Fine-tuned Models}
        \label{fig:base}
    \end{minipage}
    \hfill %
    \begin{minipage}[b]{0.56\textwidth}
        \centering
        \includegraphics[width=\textwidth, alt={A grouped bar chart displaying federated performance without synthetic data across three model sizes: 0.6B, 1.7B, and 4B. The evaluated metrics are F1-Score, MRR, and Hits@1. Performance clearly increases as the parameter size grows, with the 4B model achieving the highest scores in all three metric categories.}]{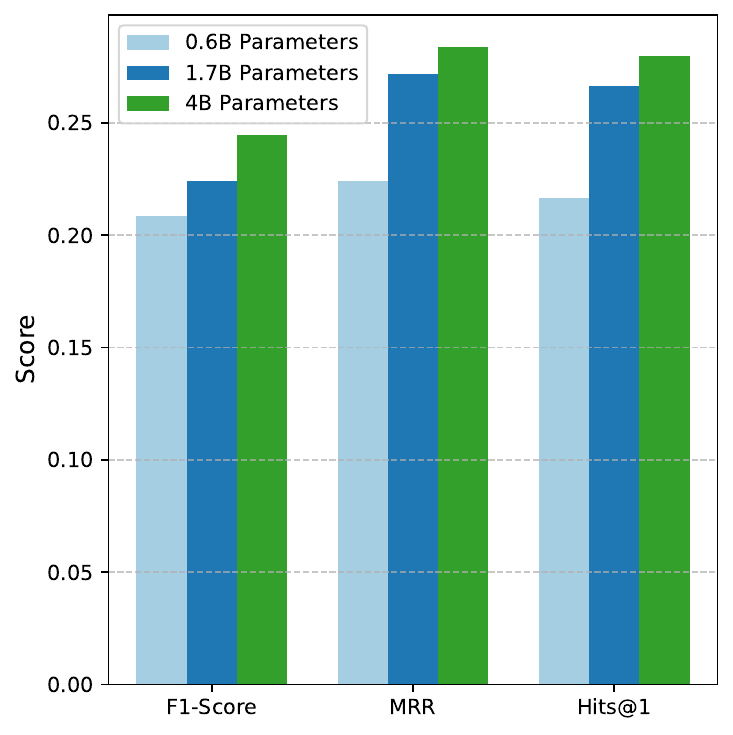}
        \caption{Federated Performance by Model Size (No Synthetic Data)}
        \label{fig:size}
    \end{minipage}
\end{figure}

\Cref{fig:base} shows that the performance of TREK-LM is improved over the base models thanks to the KTO fine-tuning. All three sizes of the Qwen3 model showed significant improvement in our experiments, with 0.6B's F1-score more than doubling.

\subsubsection{Effect of Model Size:}
As shown in \Cref{fig:size}, the performance of TREK-LM does improve when using a larger base model, but the improvement is not as drastic as one may expect. The improvement in F1-score seems to be largely linear with the number of parameters, while the results are a bit more interesting for MRR and Hits@1. There is a decent gap between 0.6B and 1.7B for both of the metrics, while 1.7B and 4B are much closer in their results, indicating there may be diminishing returns as the model size increases.

\subsubsection{Effects of Synthetic Data:}

\begin{wrapfigure}{r}{0.8\textwidth} %
    \centering
    \vspace{-10pt} %
    \includegraphics[width=\linewidth, alt={A grouped bar chart detailing the impact of synthetic data on F1-Scores for 0.6B, 1.7B, and 4B models in both Centralized and Federated settings. For the vast majority of model sizes and settings, the bars representing training with synthetic additions are lower than their non-synthetic counterparts, visually indicating a degradation in performance when synthetic data is introduced.}]{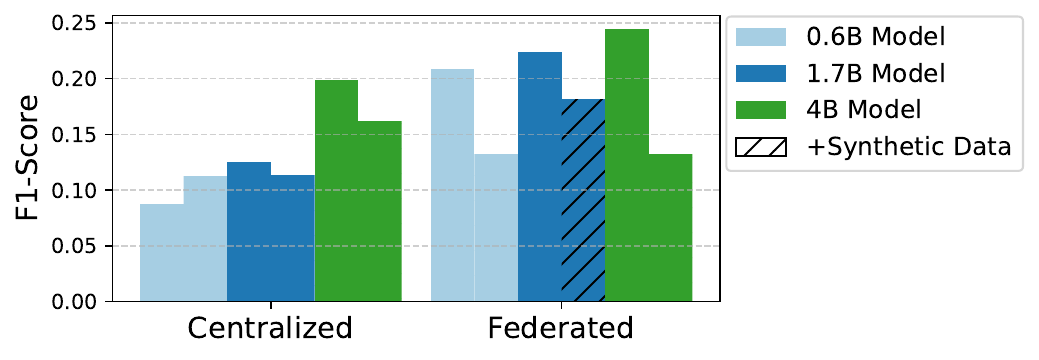}
    \vspace{-3mm}
    \caption{Impact of Synthetic Data Across All Models}
    \label{fig:syn}
    \vspace{-10pt} %
\end{wrapfigure}

Another key insight concerns the role of synthetic data. As can be seen in \Cref{fig:syn}, for the 0.6B model, adding synthetic examples slightly improves performance in the centralized setting.
However, in all other settings and model sizes, synthetic data leads to performance degradation. 

This suggests that the generated samples fail to capture the fine-grained, user-specific distributions essential for effective personalization and instead introduce noise or spurious correlations. In other words, synthetic data lacks the semantic and contextual fidelity of real user interactions, which limits its utility for KG completion under realistic conditions.
This underscores the value of FedTREK-LM, which makes it possible to leverage real, privacy-sensitive user data in a federated manner without compromising user sovereignty.

\subsection{Recipe PKG Results}
\label{sec:recres}

\begin{wraptable}[13]{r}{0.7\textwidth}
\vspace{-7mm}
\centering
\caption{Performance on the Recipe PKG recommendation task. All TREK-LM variants and baselines achieved near-zero performance across all metrics.}
\label{tab:recipe}
\begin{tabular}{llccc}
\toprule
\textbf{Model} & \textbf{Setting} & \textbf{F1-score} & \textbf{MRR} & \textbf{Hits@10} \\
\midrule
\multirow{3}{*}{TREK-LM} 
 & Centralized & 0.000 & 0.000 & 0.000 \\
 & Federated & 0.002 & 0.002 & 0.000 \\
 & Local & 0.000 & 0.000 & 0.000 \\
\midrule
KBGAT & Cen./Fed. & < 0.004 & < 0.017 & < 0.019 \\
HAKE & Cen./Fed. & < 0.002 & < 0.003 & < 0.007 \\
FedKGRec & Federated & 0.000 & 0.000 & 0.000 \\
\bottomrule
\end{tabular}
\end{wraptable}

In \Cref{tab:recipe}, we show the results on the Recipe PKG data. The results here were near-zero across the board for both TREK-LM and the baselines. FedKGRec once again could not handle the sparsity of the data. 
A key reason for this outcome is that the Recipe PKG dataset is constructed entirely from synthetic preference interactions, as the original Food.com data does not contain conversational signals or temporally grounded user intent.

Together, these results highlight the importance of real conversation and recommendation data for training these models, and reinforce the need for federated access to real user interactions when performing PKG-based recommendation.

\section{Conclusion}
\label{sec:conc}

In this work, we presented a novel approach to personalized user recommendation that combines a lightweight LLM with KTO in an FL setting. 
Our methodology, FedTREK-LM, supports continual learning over evolving PKGs, adapts to new entities and relations, and performs domain-specific reasoning through natural language prompts. 
It is robust to sparse data and aligns well with real-world scenarios where user intent is expressed in natural language and captured through evolving structured triples in a user's PKG.
Our findings highlight two central insights: (1) FedTREK-LM consistently outperforms existing KG completion baselines under both centralized, federated, and local setups, and (2) real user interactions, when accessed through FL, are essential for meaningful PKG completion, as they enable cross-user generalization that purely local models cannot achieve.

State-of-the-art KG completion baselines KBGAT and HAKE which assume centralized access to global KGs, degrade sharply when deployed in federated settings,
whereas FedTREK-LM maintains strong performance while respecting decentralization and scalability constraints.
Additionally, FedKGRec, despite being designed for federation, struggles with the sparsity of the datasets and fails to learn effective representations, underscoring the advantages of LLM-based PKG reasoning in sparse data regimes.

Notably, our results demonstrate that even small, quantized models such as Qwen3-0.6B, 1.7B, and 4B can achieve strong performance, making our method feasible for on-device user-centric applications. 
Additionally, further gains could be achieved through improved prompt engineering and hyperparameter tuning, particularly in adapting system prompts and user queries to maximize relevance and generalization for improved recommendation accuracy and better adaptation to domain-specific KGs.

Although our evaluation focused on the movie and food domains, the framework generalizes naturally to other PKG tasks in domains like healthcare, education, and personal finance. 
By treating PKGs as evolving user models and enabling federated, feedback-driven adaptation, FedTREK-LM advances the broader vision of user-aligned AI assistants that continuously learn and reason across decentralized environments.
Overall, FedTREK-LM offers a scalable and versatile paradigm for LLM-powered personalization, merging semantic reasoning, federated optimization, and user modeling into a unified framework. 

\subsection*{Supplementary Material Statement}
\label{sec:resources}

All research artifacts, including source code, dataset construction scripts, sample data, and result generation pipelines, are available in our GitHub repository. All external datasets and software dependencies used in this work are documented and linked in the repository’s README.\\
\url{https://github.com/brains-group/TREK-LM}. 

\bibliographystyle{splncs04}
\footnotesize{\bibliography{references}}

\newpage
\appendix
\renewcommand{\thesection}{Appendix \Alph{section}}

\section{Hardware and Software Specifications}

Our experiments were run in Ubuntu 24.04.2 LTS with an NVIDIA H100 NVL GPU with 94GB of VRAM, an Intel Xeon Gold 6448Y CPU, and 512 GB of RAM. We use the Transformer Reinforcement Learning (TRL) implementation of KTO within KTOTrainer~\cite{huggingface2024kto}. We simulate FL using the Flower framework~\cite{beutel2020flower} with 128 communication rounds, using 4 selected clients for training in each round.

\section{Hyperparameters}

\subsection{KTO Hyperparameters}  

The KTO guidelines state that the \texttt{desirable\_weight} and \texttt{undesirable\_weight} hyperparameters should have a ratio between 3:4 and 4:3~\cite{huggingface2024kto}, so we chose a 4:3 ratio to emphasize successful completions during training. 
\oshani{This choice of 4:3 ratio says nothing about why we didn't use 3:4. Can you please elaborate? Or is this supposed to be the same ratio somehow?  Let's move to appendix}
\fernando{Ok, I'm moving this to the appendix. I also added a small clarification in the next sentence to make this a little more understandable: (positive feedback is weighted higher than negative feedback)}
This weighting scheme biases the model toward learning from positive examples (positive feedback is weighted higher than negative feedback), encouraging it to generalize patterns that yield high-quality, structured outputs, such as well-formed triples and multiple relevant recommendations. At the same time, the presence of negatively weighted examples ensures that the model still learns to suppress negative recommendations. This balance is critical for steering the model toward desirable behaviors without overfitting to only idealized inputs.

\subsection{LoRA Hyperparameters}  
For our experiments, we set the LoRA rank to 16 and the scaling factor ($\alpha$) to 64. The rank controls the dimensionality of the low-rank matrices injected into the model's weight updates. Higher values offer greater expressiveness during fine-tuning but increase memory and compute overhead. The $\alpha$ parameter scales the LoRA updates relative to the base model weights, effectively amplifying their influence.

\oshani{I really think we should add the pseudocode for one round in the Appendix, especially because this community is not well-versed in FL. You can include sample clients, broadcast model, local KTO+LoRA update, send deltas to aggregator, etc. }
\fernando{There isn't enough space for this, and I don't think it would actually help anyone anyway. I edited the first paragraph of the System Architecture section a bit to describe federated learning a bit more instead.}

\section{Prompt Templates}
\label{sec:prompts}

\subsection{System Prompt}
\label{sec:system-prompt}

\begin{tcolorbox}[title=\example\label{ex:sys} The system prompt for guiding the LLM, colback=gray!5, colframe=gray!40!black]
\footnotesize
You perform \{Recommendation\_Domain\} recommendations based on a Knowledge Graph. You will recommend a list of \{Item\_Type\} to the user that are not already in their knowledge graph. The user's entity is represented by \{User\_ID\}. Use this knowledge graph when responding to their queries: \{User\_PKG\}. Provide the recommendations as a bulleted list with dashes (-) as the bullet points.
\end{tcolorbox}

\subsection{Prompt-Completion Examples}
\label{sec:pce}

\begin{tcolorbox}[title=\example\label{ex:real} Real Interaction, colback=gray!5, colframe=gray!40!black]
\footnotesize
\textbf{Prompt:}
\begin{itemize}[leftmargin=*]
\item \textbf{System}: $<\text{SYSTEM PROMPT}>$: \\\texttt{\{"32": \{"liked": ["Raiders of the Lost Ark (1981)", "Air Force One (1997)", "What Lies Beneath (2000)", "Ender's Game (2013)", "Working Girl (1988)", "The Devil's Own (1997)", "Six Days, Seven Nights (1998)", "Indiana Jones "]\}\}} 

\item \textbf{Assistant}: Hey! What actors are you interested in? I would like to help you find a good movie.

\item \textbf{User}: I really enjoy some of Harrison Ford's work. I think he was great in \textit{Air Force One (1997)} and \textit{Raiders of the Lost Ark (1981)}.
\end{itemize}

\textbf{Completion:}
\begin{itemize}
    \item \textbf{Assistant}: Based on your Knowledge Graph, I recommend the following: 

- \texttt{What Lies Beneath (2000)}
\end{itemize}
\end{tcolorbox}

\begin{tcolorbox}[title=\example\label{ex:syn} Synthetic Interaction, colback=gray!5, colframe=gray!40!black]
\footnotesize
\textbf{Prompt:}
\begin{itemize}[topsep=2pt, itemsep=2pt, parsep=0pt,leftmargin=*]
    \item \textbf{System}: $<\text{SYSTEM PROMPT}>$: \\\texttt{\{"0": \{"liked": ["Mary and Max (2009)", "The Boss Baby (2017)", "A Scanner Darkly  (2006)", "Final Fantasy: The Spirits Within (2001)"]\}\}}
    \item \textbf{User}: Recommend movies to me.
\end{itemize}
\textbf{Completion:}
\begin{itemize}
    \item \textbf{Assistant}: Based on your Knowledge Graph, I recommend the following:

- \texttt{Waking Life (2001)} \\
- \texttt{The Triplets of Belleville (2003)}
\end{itemize}
\end{tcolorbox}

\subsection*{GenAI Usage Disclosure:}
GenAI was used to help generate some parts of the code as well as the paper content. The authors retain full responsibility for the accuracy, originality, and overall content of the code and the paper.

\end{document}